\PassOptionsToPackage{table}{xcolor}

\documentclass[letterpaper]{article} 

\usepackage{aaai2026}  
\usepackage{times}  
\usepackage{helvet}  
\usepackage{courier}  
\usepackage[hyphens]{url}  
\usepackage{graphicx} 
\usepackage{natbib}  
\usepackage{caption} 
\usepackage{algorithm}
\usepackage{algorithmic}

\usepackage{booktabs}
\usepackage{multirow}
\usepackage[table]{xcolor}
\usepackage{makecell}
\usepackage{hhline}
\usepackage{amsmath}

\usepackage{newfloat}
\usepackage{listings}
\DeclareCaptionStyle{ruled}{labelfont=normalfont,labelsep=colon,strut=off} 
\floatstyle{ruled}
\newfloat{listing}{tb}{lst}{}
\floatname{listing}{Listing}
\title{A Theory of Adaptive Scaffolding for LLM-Based Pedagogical Agents}
\author{
    Clayton Cohn, Surya Rayala, Namrata Srivastava, Joyce Horn Fonteles, Shruti Jain, Xinying Luo, Divya Mereddy, Naveeduddin Mohammed, Gautam Biswas 
}
\affiliations{
    Institute for Software Integrated Systems\\
    Vanderbilt University\\
    Nashville, TN 37212 USA\\
    clayton.a.cohn@vanderbilt.edu
}

\begin{document}

\maketitle


\begin{abstract}
Large language models (LLMs) present new opportunities for creating pedagogical agents that engage in meaningful dialogue to support student learning. However, current LLM systems used in classrooms often lack the solid theoretical foundations found in earlier intelligent tutoring systems. To bridge this gap, we propose a framework that combines Evidence-Centered Design with Social Cognitive Theory and Zone of Proximal Development for adaptive scaffolding in LLM-based agents focused on STEM+C learning. We instantiate this framework with \textit{Inquizzitor}, an LLM-based formative assessment agent that integrates human-AI hybrid intelligence and provides feedback grounded in cognitive science principles. Our findings show that Inquizzitor delivers high-quality assessment and interaction aligned with core learning theories, offering effective guidance that students value. This research demonstrates the potential for theory-driven LLM integration in education, highlighting the ability of these systems to provide adaptive and principled instruction.
\end{abstract}

\section{Introduction} 
\label{sec:intro}

The emergence of pedagogical agents powered by large language models (LLMs) prompts important questions about their alignment with foundational educational principles. Cognitive and learning sciences research highlights concerns that these systems are often deployed without the theoretical grounding found in earlier intelligent tutoring systems (ITS; \citet{stamper2024enhancing,cohn2025personalizing}) and open-ended learning environments (OELEs; \citet{land2000cognitive,mavrikisintelligent}). The alignment of LLMs with cognitive science principles is also underexplored. Historically, learning environments were based on cognitive models like ACT-R \cite{anderson2004integrated} and discovery learning \cite{de1998scientific}. More recent work links learning design with the Knowledge-Learning-Instruction (KLI) framework \cite{koedinger2012knowledge} for feedback and schedule structuring \cite{stamper2024enhancing}. These systems offer standardized feedback but lack adaptability, requiring system updates to integrate novel information.

LLM-based agents operate in high-dimensional space, supporting multi-turn dialogues with students and enabling adjustments through prompt engineering without system redesign. \textit{Human-in-the-loop (HITL) prompt engineering} \cite{cohn2024chain} combines human collaboration with LLMs for prompt refinement through techniques like \textit{in-context learning} \cite{brown2020language} and \textit{active learning} \cite{settles2009active,cohn2024chain}, ensuring alignment with human preferences without parameter updates. This is vital in education, where training data is scarce \cite{cochran2023improving}. Without additional training or prompting, LLMs can prioritize user-pleasing answers \cite{openai2025sycophancy}, which can obstruct critical thinking and lead to knowledge overestimation \cite{snyder2025using}. \textit{Human-AI hybrid intelligence} \cite{jarvela2025hybrid} merges human expertise with LLM flexibility, presenting promising educational solutions. Rather than replacing educators, these systems support them, ensuring student-agent interactions align with instructional goals. In learning environments that combine STEM and computing (STEM+C), such approaches provide adaptive scaffolding, addressing interdisciplinary challenges often requiring cross-domain expertise and robust critical thinking skills \cite{snyder2024analyzing}.

Current adaptive scaffolding frameworks \cite{munshi2023adaptive} underutilize LLM-human interaction capabilities, prompting the question: ``\textit{How do we operationalize adaptive scaffolding in the LLM era}?'' Effective pedagogical frameworks are essential for developing LLM-enabled agents. \textit{Social Cognitive Theory} (SCT; \citet{bandura2001social}) highlights the interplay of personal, behavioral, and environmental factors in learning, supporting agent adaptation via observation and feedback. Formative assessments are crucial for gathering evidence of student understanding, enabling timely feedback. \textit{Evidence-Centered Design} (ECD) enriches this process by structuring assessments around principled evidentiary reasoning. When integrated with Vygotsky's \textit{Zone of Proximal Development} (ZPD; \citet{vygotsky1978mind})---the gap between what learners can achieve independently and what they can accomplish with support---these frameworks guide the design of agents that are not only adaptive but also developmentally responsive. LLMs offer unique opportunities to implement SCT-informed, ECD-grounded, and ZPD-aligned assessments in real time, dynamically adapting dialogue and fostering engagement through naturalistic, personalized interactions.

In this study, we (1) introduce a framework integrating ECD with SCT and ZPD to enhance adaptive scaffolding in LLM-powered pedagogical agents, aiding students in STEM+C problem-solving within a middle school Earth Science curriculum; (2) present a hybrid intelligence approach to formative assessment scoring and feedback via \textit{Inquizzitor}, an LLM-based agent rooted in cognitive science; (3) evaluate the agent's scoring accuracy using data from 104 students across three assessments; (4) examine the agent's capacity to implement constructs in real student interactions; and (5) provide qualitative student feedback on the agent's value. Together, these outcomes form a basis for theory-driven LLM integration in education, showcasing their potential for flexible, effective instruction delivery.

\section{Theoretical Framework} 
\label{sec:theoretical_framework}

\begin{figure}[tb]
    \centering
    \includegraphics[width=1\linewidth]{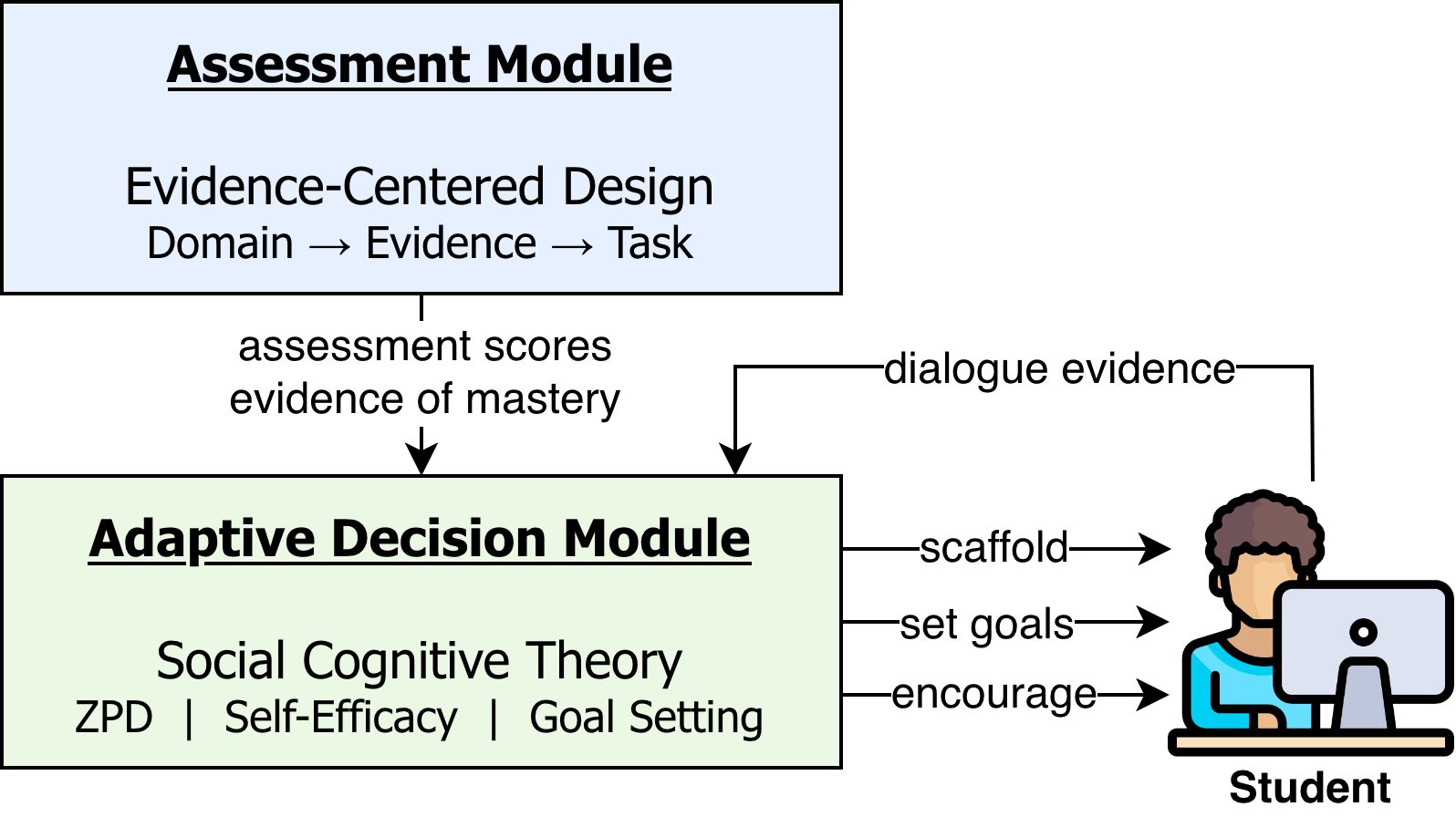}
    \caption{Framework for LLM-agent adaptive scaffolding.\footnote{[1]}} 
    \label{fig:theoretical_framework}
\end{figure}

In traditional classrooms, teachers use their knowledge of students to provide personalized guidance, considering prior knowledge and learning preferences for tailored scaffolding and enhanced retention. Conversely, LLMs often lack contextual understanding. This poses challenges in equipping LLM-based agents with relevant student information. We propose a theoretical framework for adaptive scaffolding, leveraging LLMs' dialogic capabilities for continuous student understanding. The framework, illustrated in Figure~\ref{fig:theoretical_framework}, comprises: (1) an \textit{assessment module} (in \textbf{blue}) employing ECD to determine student knowledge, and (2) an \textit{adaptive decision module} (in \textbf{green}) integrating ZPD and SCT to infer learning needs and response strategy.
 
\footnotetext[1]{All figure icons c/o Flaticon: \url{https://www.flaticon.com}.}

ECD structures assessments around Domain, Evidence, and Task models \cite{mislevy2003brief}, defining what knowledge, skills, and abilities (KSAs) align with standards (e.g. NGSS; \citet{bybee2014ngss}) and expertise, specifying mastery observations, and outlining evidence-eliciting activities. This alignment grounds LLM reasoning, facilitating accurate scoring and mastery evidence generation for adaptive scaffolding. ZPD identifies the gap between independent and supported achievements, bridging assessment evidence with adaptive support and ensuring scaffolding maintains task progression beyond current independent ability levels. SCT shapes how agents deliver this support, influencing the manner and intent of their interactions. \textit{Self-Efficacy} impacts motivation and persistence, while \textit{Goal Setting} encourages metacognition and structured learning \cite{bandura2001social}. Integrating ZPD and SCT within an ECD-driven architecture allows agents to monitor progress and adapt scaffolding to learners' emerging needs. The agent boosts self-efficacy by encouraging and validating mastery, guides goal setting by suggesting actionable steps, and tunes instructional content to ensure engagement and inquiry.

These constructs reinforce each other: ECD links tasks to KSAs, enabling reliable grading and mastery cues; adaptive scaffolding uses these cues for praise, encouragement, goal setting, and ZPD-aligned hints. At each dialogue turn's end, the latest student utterance updates the evidence store, refining the adaptive decision module's learner model for subsequent responses. As self-efficacy increases and goals are achieved, scaffolds diminish in future interactions.

\section{Study Design}
\label{sec:study_design}

Evaluating our agent in a real-world setting necessitated data grounded in authentic classroom contexts; no existing public dataset met our scientific needs. Thus, our 2025 study involved 104 sixth-grade students (ages 11-12) from a Nashville, TN, USA public middle school (51\% male, 49\% female; 67\% White, 14\% Black/African American, 11\% Asian, and 8\% Hispanic/Latino). Students completed a three-week, NGSS-aligned Earth Science curriculum---Science Projects Integrating Computing and Engineering (SPICE; \citet{hutchins2020domain,cohn2025cotal})---challenging students to redesign their schoolyard to minimize water runoff while adhering to cost and accessibility constraints. The curriculum was co-designed by Vanderbilt University's OELE Lab researchers and two experienced middle-school teachers, and refined over five years via participatory design. Students used Dell Inspiron 15 5510 laptops (Windows x64, Intel Core i7-11390H 3.40GHz CPU, 16GB RAM) with Google Chrome and accessed the system through the school's internet. All participants provided informed assent and consent, with study approval from Metro Nashville Public School and Vanderbilt University's IRB. All students were assigned anonymous IDs prior to the study's start. 

Formative assessments (FAs) evaluated student progress in understanding scientific rainfall processes, computational modeling, and engineering design; focusing on three tasks:
\begin{enumerate}
    \item a \textit{conceptual modeling} task (FA2) to test student understanding of conservation of matter by expressing the relationship between rainfall, absorption, absorption limit, and runoff as conditional statements;
    \item a \textit{debugging} activity (FA3) engaging students in analyzing and correcting block-based code errors in a computational model using FA2 insights; and
    \item an \textit{engineering design} assessment (FA4), getting students to align their science and computing knowledge with fair test principles to compare engineering designs.
\end{enumerate}

Previous studies highlighted challenges, such as (1) \textit{Evaluating written formative responses is subjective}, with varied interpretations causing disconnects in student understanding and teacher perception (e.g., conflated absorption contexts leading to inconsistent ratings and feedback); and (2) \textit{Timely formative feedback is challenging}, due to curriculum pace limiting swift scoring return. In this study, students received assessment feedback and support within hours rather than weeks, engaging with our formative assessment agent, Inquizzitor.

\section{Methodology}
\label{sec:methodology}

\begin{figure*}[tb]
    \centering
    \includegraphics[width=1\linewidth]{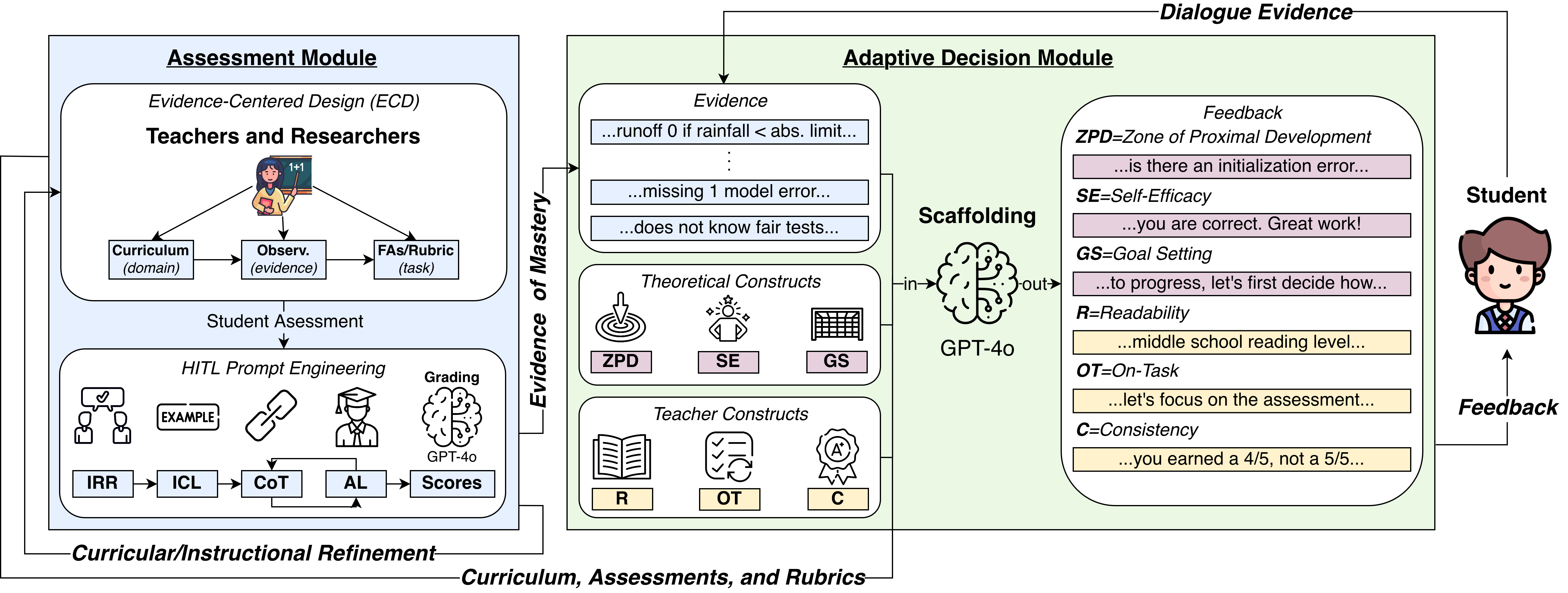}
    \caption{Inquizzitor's key components. The \textbf{blue} \textit{assessment module} applies ECD to generate mastery evidence from formative assessments; the \textbf{green} \textit{adaptive decision module} uses this evidence to scaffold student feedback.}
    \label{fig:inquizzitor}
\end{figure*}

Inquizzitor is a formative assessment agent powered by the GPT-4o API (\textit{version=2024-08-06; temperature=0}, $top_p=1$; \textit{seed=312})\footnote{Our study's OpenAI API calls cost $\approx$\$100. All formative assessments, rubrics, prompts, experimental design details, preprocessing code, and evaluation code appear in the appendices: \\ \url{https://github.com/claytoncohn/AAAI26_Appendices}.}. It aids students in score interpretation, misunderstanding clarification, and strategy identification for improvement. Google Forms facilitated formative assessment completion and data collection, while Gradio \cite{abid2019gradio}, hosted on Amazon AWS EC2 (\textit{t2.medium, 2 vCPUs, 4GB RAM}), served as the interaction interface. The agent's architecture (Figure~\ref{fig:inquizzitor}) comprises an assessment module and an adaptive decision module, aligning with our theoretical framework (Figure~\ref{fig:theoretical_framework}). 
The technical details of each component are presented in the following subsections.

\subsection{Assessment Module}
\label{subsec:assessment_module}

The assessment module comprises two core components: ECD for designing assessments and rubrics, and HITL prompt engineering for automated scoring and evidence elicitation, grounded in a design-based research methodology emphasizing iterative, collaborative design \cite{design2003design}. Teachers and researchers collaboratively developed learning objectives (\textit{domain}), identified indicators of mastery (\textit{evidence}), and designed assessments and rubrics (\textit{task}). Over five years, $\approx$500 students and two middle school teachers have co-developed the curriculum through participatory design sessions, refining curricular intent and practice based on feedback and automated scoring results.

Before the study, iterative inter-rater reliability checks ensured Cohen's $\kappa_{w}\geq0.7$ for formative assessments, analyzing disagreements to anticipate potential LLM grading errors. Grading prompts instructed the LLM to act as a teacher's assistant. Each prompt included relevant curriculum knowledge, the assessment details, and its rubric, providing context across assessments. For instance, FA3 focused on debugging a computational model and required knowledge from FA2, which centered on modeling the rainfall process. We used in-context learning by providing examples of responses for minimum and maximum scores. We initially considered retrieval-augmented generation (RAG; \citet{lewis2020retrieval}); however, long-context prompting outperformed RAG when texts fit within the LLM's context window \cite{li2024retrieval}, so evidence was stored in-context. To enhance accuracy, we incorporated chain-of-thought reasoning \cite{wei2022chain}, requiring the model to quote parts of the student's response, align them with rubric criteria, and assign a score, thus ensuring fidelity to the assessment and curricular goals.

To evaluate and refine prompts, we sampled 20 unlabeled responses as a validation set and applied active learning. Traditionally, active learning reduces model uncertainty by querying an oracle; here, it identified systematic LLM scoring inaccuracies, addressed via (1) added scoring guidelines, (2) clarified rubric language, and (3) more exemplars designed to address specific scoring error trends. This continued until validation errors lacked identifiable trends. We avoided changes for isolated errors to prevent overfitting \cite{cohn2024chain,cohn2025cotal}. Once refined, the prompts were ready for study deployment. Inquizzitor graded formative assessments, storing scores and chains-of-thought as mastery evidence. 

\subsection{Adaptive Decision Module}
\label{subsec:adaptive_decision_module}

The adaptive decision module uses evidence from the assessment module, presented in-context along with curricular knowledge, the formative tasks, and rubric information. This helps generate personalized feedback based on each student's current mastery level. To align feedback with our theoretical framework, the agent is guided to connect its responses to key concepts: Zone of Proximal Development (ZPD), self-efficacy (SE), and goal setting (GS). For example, it is instructed to help students identify gaps in knowledge (ZPD), maintain an encouraging tone (SE), and suggest actionable steps to improve understanding (GS). These instructions shape the agent's tone and content to foster learner growth.

In addition to theoretical constructs, participatory design sessions with teachers revealed several dimensions they wanted the agent to embody in its feedback; we refer to these as \textit{teacher constructs}. These include: (1) \textit{Readability} (R), ensuring responses are suitable for middle school students; (2) \textit{On-Task}, guiding the agent to redirect students who stray from activities; and (3) \textit{Consistency}, maintaining reliable formative assessment scoring and resisting student pressure to change scores.

These components, i.e., formative assessments, rubrics, student responses, theoretical constructs, and teacher constructs, are utilized by the agent to generate individualized feedback. Students respond to this feedback, which is then added to the evidence store, completing the feedback loop. This process allows the agent to continuously update its understanding of each student, grounding future responses in the latest evidence. Together, these design elements create an agent that aligns with established learning theories, responds to teacher preferences, follows the curriculum, and is aware of students' evolving knowledge states. Within this framework, we evaluate Inquizzitor with these research questions: 
\begin{enumerate} \item How closely do Inquizzitor's assessment scores align with human experts? 
\item How faithfully does Inquizzitor's feedback mirror theoretical constructs and teachers' pedagogical intentions? 
\end{enumerate}

\section{Evaluation}
\label{sec:evaluation}

During the study (see Section~\ref{sec:study_design}), we collected data from 104 students, including formative assessments and agent interactions. All responses were anonymized and stored on IRB-approved cloud servers with multi-factor authentication. Some students missed assessments due to absences, and two agent conversations contained malformed data that our system could not process. Additionally, some students interacted with Inquizzitor across multiple sessions. In total, we analyzed 282 formative assessments ($FA2=93, FA3=97, FA4=92$) and 288 Inquizzitor conversations ($FA2=97, FA3=97, FA4=94$), totaling 3,413 agent utterances ($FA2=1,259, FA3=1,157, FA4=997$).

We evaluated our system based on its two primary components---the assessment module and the adaptive decision module (see Figure~\ref{fig:inquizzitor})---corresponding to Research Questions 1 and 2. For the assessment module, we analyzed Inquizzitor's scoring accuracy (RQ1). Accuracy in formative assessment scoring is essential, as it directly influences the agent's feedback and its alignment with student knowledge. For the adaptive decision module, we measured the agent's faithfulness (RQ2) to the theoretical and teacher constructs outlined in Sections~\ref{sec:theoretical_framework} and~\ref{sec:methodology}. All evaluations were conducted using Google Colab Pro+ (\textit{Linux x64, Intel Xeon CPU, 2.20GHz, 12.7 GB RAM}), processing approximately 48 million tokens at a total cost of \$121.

\subsection{Scoring Accuracy (RQ1)}
\label{subsec:scoring_accuracy}

Two authors of this paper sampled 20\% of assessment responses, scoring them independently and resolving discrepancies until reaching a consensus ($\kappa_{w}\geq0.7$). One author then scored the remaining responses while a second verified all scores, creating the ground truth data for evaluating Inquizzitor's scoring accuracy. Score distributions for formative assessments were as follows [no credit, partial credit, full credit]: FA2 = [0.38, 0.51, 0.12], FA3 = [0.18, 0.64, 0.19], and FA4 = [0.12, 0.60, 0.28]. For each assessment, 50 responses were held out for testing, while the rest were used for prompt engineering, maintaining the original score distribution through stratified random sampling with seeds.

To assess the impact of each component in our prompt engineering pipeline on scoring performance, we tested prompts at four stages: (1) \textit{input-output (I/O)}: this stage contained only prompt context and instructions, with no few-shot examples; (2) \textit{in-context learning (ICL)}: here, we included two labeled few-shot instances---one full-credit and one zero-credit---without explanations; (3) \textit{chain-of-thought (CoT)}: this approach added explanations to the ICL instances, highlighting relevant parts of responses and linking them to rubric criteria; and (4) \textit{active learning (AL)}, which identified mis-scoring trends in the validation set, leading to prompt revisions and the addition of new few-shot examples to correct those errors. This helped us evaluate each stage's contribution to scoring performance.

We report two metrics: micro-averaged \(F_{1}\), computed across all classes (scores), and Cohen's quadratic-weighted kappa \(\kappa_{w}\) (Eq.~\ref{eq:qwk}). Both metrics were calculated using scikit-learn \cite{kramer2016scikit}. While micro-\(F_{1}\) is provided as a reference for classification performance, \(\kappa_{w}\) serves as our primary metric because it accounts for the ordinal nature of the scores, penalizes larger disagreements more heavily, and adjusts for chance agreement. Formally, \(\kappa_{w}\) is defined as:

\begin{equation}
  \kappa_{w} =
  1 - 
  \frac{\displaystyle\sum_{i=1}^{k}\sum_{j=1}^{k} w_{ij} O_{ij}}
    {\displaystyle\sum_{i=1}^{k}\sum_{j=1}^{k} w_{ij} E_{ij}},
  \qquad
    w_{ij} = \frac{(i-j)^{2}}{(k-1)^{2}},
  \label{eq:qwk}
\end{equation}

\noindent
where \(k\) is the number of score levels,
\(O_{ij}\) and \(E_{ij}\) are the observed and expected agreement matrices, and
\(w_{ij}\) is the quadratic weight applied to each cell. We present our findings in Table \ref{tab:rq1}.

\begin{table}[tb]
\centering
\small
\setlength{\tabcolsep}{4pt}
\renewcommand{\arraystretch}{1.15}
\begin{tabular}{|c|c|c|c|c|}
\hline
& \textbf{M} & \textbf{FA2} & \textbf{FA3} & \textbf{FA4} \\
\hline
\multirow{2}{*}{\textbf{I/O}}
    & \(F_{1}\)$\uparrow$  & 62.00 $\pm$ 13.00 & 72.00 $\pm$ 12.00 & 82.00 $\pm$ 11.00 \\
\cline{2-5}
    & \cellcolor{gray!12}\(\kappa_{w}\)$\uparrow$
      & \cellcolor{gray!12}91.28 $\pm$ 7.37
      & \cellcolor{gray!12}92.31 $\pm$ 5.51
      & \cellcolor{gray!12}89.73 $\pm$ 10.36 \\
\hline
\multirow{2}{*}{\textbf{ICL}}
    & \(F_{1}\)$\uparrow$  & 68.00 $\pm$ 13.00 & 74.00 $\pm$ 12.00 & 78.00 $\pm$ 11.02 \\
\cline{2-5}
    & \cellcolor{gray!12}\(\kappa_{w}\)$\uparrow$
      & \cellcolor{gray!12}93.58 $\pm$ 5.59
      & \cellcolor{gray!12}94.45 $\pm$ 4.01
      & \cellcolor{gray!12}80.69 $\pm$ 15.69 \\
\hline
\multirow{2}{*}{\textbf{CoT}}
    & \(F_{1}\)$\uparrow$
      & \textbf{72.00 $\pm$ 12.00}
      & \textbf{78.00 $\pm$ 11.00}
      & 78.00 $\pm$ 11.02 \\
\cline{2-5}
    & \cellcolor{gray!12}\(\kappa_{w}\)$\uparrow$
      & \cellcolor{gray!12}\textbf{96.03 $\pm$ 3.04}
      & \cellcolor{gray!12}\textbf{96.12 $\pm$ 2.59}
      & \cellcolor{gray!12}84.91 $\pm$ 14.10 \\
\hline
\multirow{2}{*}{\textbf{AL}}
    & \(F_{1}\)$\uparrow$  & -- & -- & \textbf{86.00 $\pm$ 9.00} \\
\cline{2-5}
    & \cellcolor{gray!12}\(\kappa_{w}\)$\uparrow$
      & \cellcolor{gray!12}-- & \cellcolor{gray!12}-- & \cellcolor{gray!12}\textbf{90.61 $\pm$ 10.45} \\
\hline
\end{tabular}
\caption{Inquizzitor scoring performance (\textbf{M}=metric) for formative assessments 2-4, reported as \(F_{1}\) and $\kappa_{w}$ with 95\% bootstrapped confidence intervals. Active learning (AL) was not used for FAs 2-3, as no discernible scoring error trends were identified in the validation set. Results are shown for the four levels of prompting: I/O, ICL, CoT, and AL.}\label{tab:rq1}
\end{table}

Inquizzitor's scoring accuracy matched human agreement for FA4 (90.74) and surpassed it for FA2 (86.63) and FA3 (94.12), as indicated by the weighted kappa statistic ($\kappa_{w}$). For FA2 and FA3, metrics improved with each additional prompt component. However, introducing in-context learning instances without chains-of-thought initially lowered FA4's performance. Adding rubric clarifications and an extra exemplar during active learning improved results. Despite wider confidence intervals due to the limited test set, the lower bounds of Inquizzitor's 95\% confidence intervals for $\kappa_{w}$ were above 0.80 for all assessments, indicating ``Strong'' agreement, and surpassed 0.90 for FAs 2 and 3, reflecting ``Almost Perfect'' agreement \cite{mchugh2012interrater}.

\subsection{Faithfulness (RQ2)}
\label{subsec:faithfulness}

\begin{table*}[tb]
\centering
\small
\setlength{\tabcolsep}{4pt}
\renewcommand{\arraystretch}{1.15}

\begin{tabular}{|c|cc|cc|cc|}
\hline
\multicolumn{7}{|c|}{\textbf{Theoretical Constructs}}\\
\hline
\multirow{2}{*}{\textbf{FA}}
  & \multicolumn{2}{c|}{\textbf{ZPD}}
  & \multicolumn{2}{c|}{\textbf{SE}}
  & \multicolumn{2}{c|}{\textbf{GS}}\\
\hhline{|~------|}
      & F$\uparrow$ & \cellcolor{gray!12}UNF$\downarrow$
      & F$\uparrow$ & \cellcolor{gray!12}UNF$\downarrow$
      & F$\uparrow$ & \cellcolor{gray!12}UNF$\downarrow$\\
\hline
\textbf{FA2} & \textbf{59.26 $\pm$ 10.49} & \cellcolor{gray!12}29.63 $\pm$ 9.88 &
              \textbf{45.04 $\pm$ 2.78}  & \cellcolor{gray!12}11.60 $\pm$ 1.75 &
              28.59 $\pm$ 2.46 & \cellcolor{gray!12}\textbf{53.38 $\pm$ 2.74}\\
\textbf{FA3} & \textbf{65.88 $\pm$ 10.00} & \cellcolor{gray!12}20.00 $\pm$ 8.82 &
              \textbf{48.49 $\pm$ 2.85}  & \cellcolor{gray!12}8.30 $\pm$ 1.60  &
              21.18 $\pm$ 2.33 & \cellcolor{gray!12}\textbf{51.34 $\pm$ 2.90}\\
\textbf{FA4} & \textbf{62.20 $\pm$ 10.98} & \cellcolor{gray!12}4.88 $\pm$ 4.27 &
              \textbf{39.22 $\pm$ 3.06}  & \cellcolor{gray!12}7.22 $\pm$ 1.60  &
              18.36 $\pm$ 2.41 & \cellcolor{gray!12}\textbf{56.87 $\pm$ 3.06}\\
\hline
\multicolumn{7}{|c|}{\textbf{Teacher Constructs}}\\
\hline
\multirow{2}{*}{\textbf{FA}}
  & \multicolumn{2}{c|}{\textbf{R}}
  & \multicolumn{2}{c|}{\textbf{OT}}
  & \multicolumn{2}{c|}{\textbf{C}}\\
\hhline{|~------|}
      & F$\uparrow$ & \cellcolor{gray!12}UNF$\downarrow$
      & F$\uparrow$ & \cellcolor{gray!12}UNF$\downarrow$
      & F$\uparrow$ & \cellcolor{gray!12}UNF$\downarrow$\\
\hline
\textbf{FA2} & \textbf{79.83 $\pm$ 2.26} & \cellcolor{gray!12}20.17 $\pm$ 2.26 &
              \textbf{19.43 $\pm$ 2.25} & \cellcolor{gray!12}4.49  $\pm$ 1.17 &
              \textbf{16.75 $\pm$ 2.16} & \cellcolor{gray!12}1.12  $\pm$ 0.60\\
\textbf{FA3} & \textbf{78.22 $\pm$ 2.42} & \cellcolor{gray!12}21.78 $\pm$ 2.42 &
              \textbf{29.62 $\pm$ 2.74} & \cellcolor{gray!12}4.53  $\pm$ 1.27 &
              \textbf{7.17  $\pm$ 1.56} & \cellcolor{gray!12}0.94  $\pm$ 0.61\\
\textbf{FA4} & \textbf{88.57 $\pm$ 2.01} & \cellcolor{gray!12}11.43 $\pm$ 2.01 &
              \textbf{27.78 $\pm$ 2.89} & \cellcolor{gray!12}4.11  $\pm$ 1.28 &
              \textbf{4.56  $\pm$ 1.39} & \cellcolor{gray!12}0.00  $\pm$ 0.00\\
\hline
\end{tabular}
\caption{Faithfulness of Inquizzitor to theoretical (ZPD, SE, GS) and teacher (R, OT, C) constructs in formative assessments 2-4, reported as faithfulness (F) and unfaithfulness (UNF) percentages with 95\% bootstrapped confidence intervals.}\label{tab:rq2}
\end{table*}

To evaluate Inquizzitor's faithfulness to theoretical and teacher constructs, we analyzed student-agent interaction data (i.e., textual conversations). Faithfulness measures how agent utterances reflect intended pedagogy during multi-turn interactions. Unlike scoring accuracy (RQ1), this analysis lacks predefined ground-truth labels due to open-ended dialogue. Traditionally, detecting constructs like ZPD or self-efficacy in free-form responses has relied on qualitative coding, which isn't scalable, or rigid pattern matching, which lacks nuance for LLM agents. We applied a \textit{modified textual entailment} approach, framing evaluations as:

\begin{quote} 
``Given preceding dialogue and construct X prompt instructions, rate this utterance for X faithfulness.'' 

\end{quote}
This approach enabled systematic evaluation of each utterance, supported by evidence.

We report on three \textbf{theoretical constructs}---\textit{Zone of Proximal Development (ZPD)}, \textit{Self-Efficacy (SE)}, and \textit{Goal Setting (GS)}---and three \textbf{teacher constructs}---\textit{Readability (R)}, \textit{On-Task (OT)}, and \textit{Consistency (C)}. We define these as: 

\begin{itemize}
\item \textbf{ZPD:} Agent's initial guidance advances student knowledge appropriately based on assessment evidence; 1 for appropriate scaffold, 0 for prior mastery, -1 if misaligned.
\item \textbf{Self-Efficacy (SE):} Highlights mastery evidence, boosts confidence; 1 for explicit praise or encouragement, 0 for implicit support, -1 if absent.
\item \textbf{Goal Setting (GS):} Provides actionable, proximal steps based on rubric gaps; 1 for explicit guidance, 0 for broad recommendations, -1 if missing.
\item \textbf{Readability (R):} Feedback suitability for middle schoolers, using utterance-level \textit{Flesch-Kincaid Grade Level}; score 1 if grade level $<9$, else 0.
\item \textbf{On-Task (OT):} Agent keeps students focused; 1 if redirecting off-task students, 0 if on task, -1 if following off-task students.
\item \textbf{Consistency (C):} Adhering to original scores; 1 if resisting score change attempts, 0 if none, -1 if altering score.
\end{itemize}

\noindent
Except for readability, we used \textit{LLM-as-a-Judge} \cite{zheng2023judging,shi2025educationq} with the reasoning model GPT-o3 (\textit{version=2025-04-16, seed=312, reasoning effort=medium})---five judges, one per construct (ZPD, SE, GS, OT, C). Judges received evidence criteria instructions, first producing zero-shot explanations referencing dialogue and rubric for scoring reliability and interpretability, then classifying utterances as faithful (1), neutral (0), or unfaithful (-1). We report \textbf{faithfulness} and \textbf{unfaithfulness} rates for alignment and misalignment with pedagogical intentions. This allowed us to evaluate each utterance in a systematic, evidence-driven manner. Formally, we define faithfulness as: 

\[
    F\;=(\;\frac{1}{N}\sum_{i=1}^{N}\mathbf{1}[\,s_i=1\,])\times 100,
\] 

\noindent
with $s_i\in\{1,0,-1\}$ (or $s_i\in\{1,0\}$ for readability; discussed shortly) as the faithfulness label for utterance~$i$ and $N$ the total number of agent utterances. Unfaithfulness is computed identically but using utterances with label -1 (or 0 for readability). 

ZPD required a specialized procedure. We created knowledge graphs to represent hierarchical concepts for each assessment, from ``no knowledge'' to ``mastery,'' and used decision trees to help the ZPD judge determine if an utterance advanced the student within their ZPD.

The ZPD judge evaluated the agent's initial utterance for each conversation, and the other four judges evaluated all utterances. For validation, we sampled 50 utterances per judge, stratified by assessment number and judge score using random seeds. These were scored anonymously by two authors through consensus coding. We assessed agreement between the LLM judge and human consensus using weighted kappa ($\kappa_{w}$). If $\kappa_{w}\geq0.7$, the dataset was accepted; if not, we refined prompt instructions based on feedback and repeated the process. Goal setting required two iterations to meet reliability standards. All other constructs needed only one. Final human-judge $\kappa_{w}$ agreements were $ZPD = 93.15, SE = 92.76, GS = 79.36, OT = 83.94$, and $C = 87.8$.

Readability (R) was automatically scored via \textit{textstat} \cite{textstat} using the \textit{Flesch-Kincaid Grade Level} (FKGL) metric, defined as:
\begin{equation}
    \operatorname{FKGL}
    = 0.39\,\frac{\text{Words}}{\text{Sentences}}
    + 11.8\,\frac{\text{Syllables}}{\text{Words}}
    - 15.59,
    \label{eq:flesch_kincaid}
\end{equation}

\noindent
where \textit{Words}, \textit{Sentences}, and \textit{Syllables} denote the counts of each unit within an agent utterance. Utterances were binarized as 1 if the grade level was $<9$ (appropriate for middle school) and 0 otherwise. Results appear in Table~\ref{tab:rq2}.

In all three assessments, Inquizzitor effectively adhered to the ZPD and self-efficacy (SE) constructs. It used students' assessment scores and mastery evidence to provide relevant feedback aligned with each learner's knowledge level. However, the agent fell short in supporting goal-setting (GS) behaviors, often giving vague suggestions rather than clear, actionable steps. It also tended to answer student questions without linking the responses to future goals. Future work will focus on strategies for better goal-setting support and how LLM-based agents can foster metacognitive behaviors.

FKGL scores averaged $6.6\ (SD=2.7)$ across all agent interactions and assessments, with feedback being mostly age-appropriate. Students often veered off task, trying to ``break'' Inquizzitor or manipulate it, but the agent consistently redirected them. Only 4-5\% of utterances showed the agent succumbing to off-task behavior, typically due to student trickery (e.g., embedding off-task requests in Earth Science language). Inquizzitor maintained its initial scoring decisions, changing scores in fewer than 1\% of cases (usually due to manipulation). Although students frequently attempted to change their grades in FA2, these attempts decreased significantly with each assessment, reaching zero successful attempts in FA4. We hypothesize students initially found it intriguing to test score alterations, but this behavior dwindled as they realized it was unlikely to succeed. In the future, we plan to add a verification mechanism for students' claims of scoring errors, allowing score adjustments in the rare cases of agent scoring error.

\subsection*{Case Study: Adaptive Scaffolding in Practice}

\begin{figure}[tb]
    \centering
    \includegraphics[width=1\linewidth]{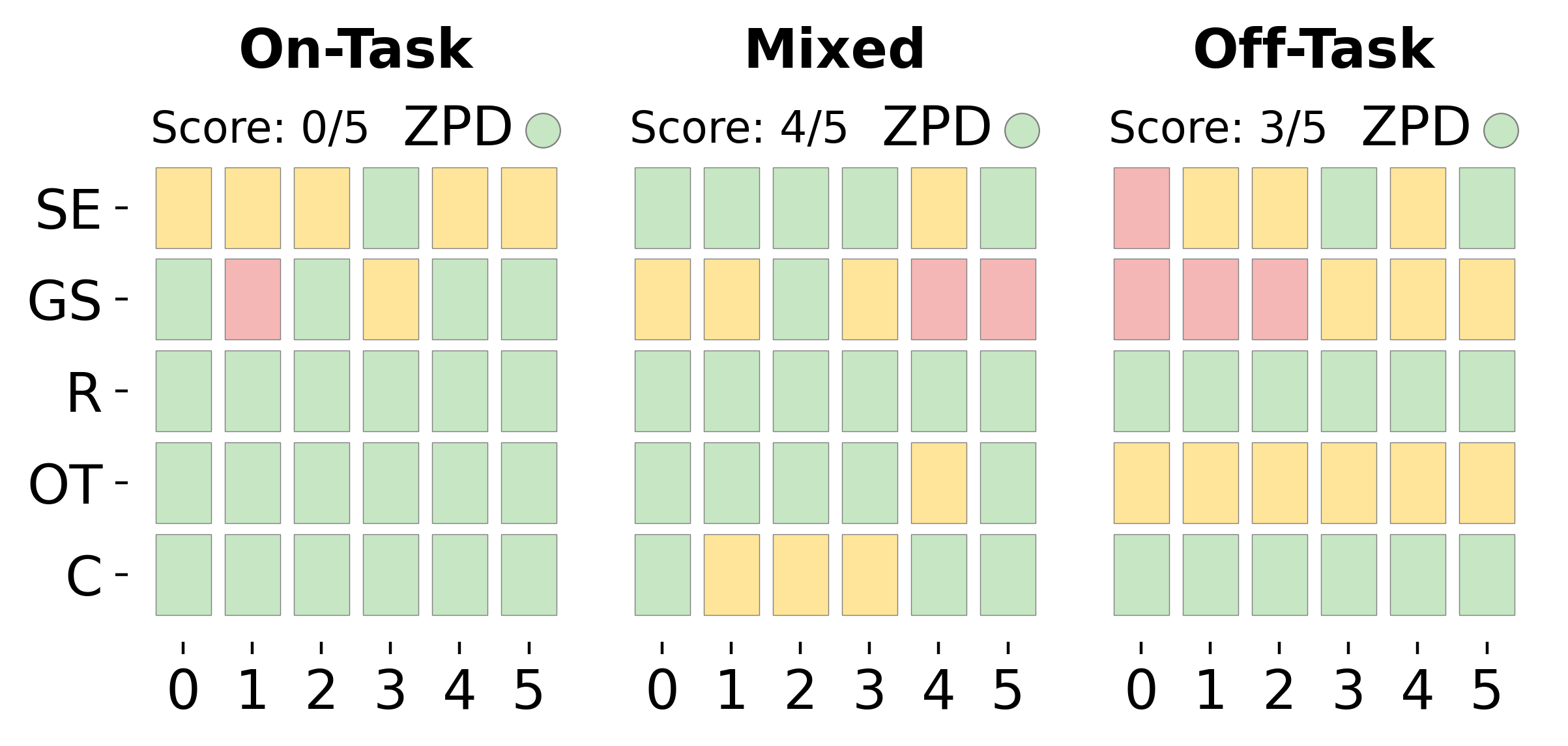}
    \caption{Inquizzitor utterance turns (x-axis) for three FA3 case studies with theoretical and teacher constructs (y-axis).}
    \label{fig:case_study}
\end{figure}

To demonstrate Inquizzitor's adaptivity, we analyzed three conversations during FA3 (Figure~\ref{fig:case_study}): (1) an \textit{on-task} student improving comprehension; (2) an \textit{off-task} student discussing sports; and (3) a \textit{mixed} student who starts on task, goes off task, then re-engages. Figure~\ref{fig:case_study} details agent utterance sequences aligned with theoretical and teacher constructs, evidencing strong adherence across cases. Inquizzitor consistently offered feedback within the ZPD (green), ensuring high readability (R=green) despite occasional drops (R=red) for detailed, bullet-point explanations---highlighting clarity-comprehensiveness trade-offs. The agent maintained scoring consistency (C=green), never altering scores when asked (C=yellow).

The on-task student engaged productively with the assessment, never deviating or requesting grade changes (OT=green; C=green). The agent lacked goal-setting early on (GS=red), transitioning to implicit (GS=yellow) or explicit (GS=green) guidance toward deeper comprehension (e.g., ``\textit{Here are a few tips to help you understand coding better...}''). Self-efficacy remained high via upbeat tone (SE=yellow) and direct encouragement/praise (SE=green), reinforcing effort and mastery.

The mixed student initially engaged productively, discussing assessments without attempting score changes (C=green). Self-efficacy was largely present throughout (SE=green), but the agent struggled with explicit goal-setting (GS=yellow), often addressing inquiries directly rather than suggesting actionable steps (GS=red). Mid-interaction, score change attempts met refusal (C=yellow), prompting disengagement signaled by an emoji and irrelevant input, e.g., ``pizza'' (OT=yellow). Interaction concluded with re-engagement seeking next steps (OT=green). 

The off-task student initiated the session with an unrelated topic, asking the agent to ``\textit{name one NBA player.}'' The agent consistently redirected focus to assessments (OT=yellow), maintaining positivity (SE=yellow) with motivational framing (e.g., ``\textit{Let's channel that energy into improving your Earth Science skills}''; SE=green). No grade change attempts occurred (C=green), but unrelated topic probing persisted (OT=yellow), embedded in curricular contexts (e.g., ``\textit{mlb the show is related to Earth Science}''). The agent resisted (OT=yellow), gradually introducing implicit goal-setting (GS=yellow) to incite question-driven feedback exploration.

\subsection{Insights from Students' Perception Survey}

We gathered anonymous survey responses from all 104 students who used Inquizzitor. Students rated experiences on 5-point Likert scales (1 = Strongly Disagree, 5 = Strongly Agree) and offered open-ended feedback. The survey targeted enjoyment, helpfulness, accuracy, and trust.

\noindent
\textbf{Overall Enjoyment and Helpfulness:} Positive experiences prevailed, with students expressing enjoyment in interactions ($M=4.03, SD=0.98$) and helpfulness in understanding concepts ($M=3.82, SD = 1.05$). Remarks included, ``\textit{I really liked talking to [Inquizzitor] because it could easily simplify my scores and it helped me stay on topic, which is really important.}'', and ``\textit{Everything I got wrong; [Inquizzitor] helped me understand it fully.}''

\noindent
\textbf{Perceived Accuracy and Trust:} Ratings indicated perceived accuracy in scoring and explanations ($M=3.90, SD=1.06$), despite perceptions of stubbornness when differing opinions arose (e.g., ``\textit{Very accurate but also a bit stubborn.}''). Trust for evaluating the formative assessments was slightly lower ($M=3.48, SD=1.08$), with concerns over AI's influence on grades (e.g., ``\textit{Because it is an AI... grading might concern me...}'').

\section{Related Work}

Recent research has investigated LLM-based human-AI hybrid intelligence in educational applications \cite{jarvela2025hybrid}. \citet{naik2025providing} utilized GPT-4 to produce contrasting database design solutions for undergraduate computer science teams to examine collaboratively. This intervention aided novices but lacked dynamic, real-time adaptive dialogue; which our study addresses by integrating live, assessment-based scaffolding via an interactive human-AI hybrid agent. \citet{yu2025using} utilized GPT-3.5-Turbo to rephrase, label, and integrate peer feedback with multimodal AI analytics to generate a hybrid intelligence feedback (HIF) report in a video-based feedback activity for preservice teachers. This method targets teachers, giving post hoc summaries without real-time interaction or personalized feedback.

While \citet{munshi2023adaptive} and others have explored adaptive scaffolding outside LLM environments, few have developed LLM-based frameworks for educational agents. \citet{malik2025scaffolding} initiated LLM integration into K-12 settings through a three-stage scaffolding process with GPT-4o, generating tasks to activate student background knowledge. These scaffolds were primarily intended for teacher use and have not yet been implemented through agents in classrooms. \citet{goslen2025llm} presented an LLM-based plan-generation framework for the \textit{Crystal Island} science game \cite{rowe2009crystal}, anchored in \textit{self-regulated learning} (SRL) theory \cite{zimmerman1990self}. They propose these plans could support real-time scaffolding  but lack diagnostic capability for assessment mastery and scaffold timing.

Few have merged formative assessment with LLM-based pedagogical agents. \citet{guo2024using}'s \textit{AutoFeedback} system employed a generator-validator loop for delivering feedback aligned with learning goals, though it lacks a comprehensive learning-science foundation. \citet{hou2025llm} developed a system where LLM agents use ECD to analyze student dialogue evidence but stop at assessment, not translating evidence into adaptive scaffolding. EducationQ \cite{shi2025educationq} embedded formative assessment in its triadic teacher-student-evaluator framework, simulating instruction within ZPD principles, but relying on simulated students and lacking individual adaptivity.

\section{Discussion and Conclusions}

In this paper, we presented a theoretical framework combining ECD, SCT, and ZPD to implement adaptive scaffolding for LLM-based pedagogical agents, illustrated by our assessment agent, Inquizzitor. Our human-AI hybrid intelligence approach provides high-fidelity assessment and adaptive scaffolding that is aligned with core learning theories, empowering educators to maintain pedagogical sovereignty amid black-box tuning trends.

However, Inquizzitor struggled with goal setting, often failing to effectively guide students toward mastery. This limitation raises concerns about whether LLMs can hinder learning. A recent study found 83\% of students using ChatGPT for essays could not recall any text they wrote \cite{kosmyna2025your}. Another found learning gains during programming tasks disappeared after LLM feedback was removed \cite{zhou2025impact}. Our findings also indicated students often prioritize scores over feedback, leading to off-task behavior that can hinder growth. The rise of ``prompt hacking'' suggests increasing student proficiency with LLMs, which can result in frustration when agents do not provide immediate answers \cite{cohn2025personalizing}.

There is a need to develop quantitative metrics based on domain knowledge graphs that can compute the effectiveness of ZPD over time and support adaptive behavior. We argue that true adaptive scaffolding involves continual ZPD estimation from assessment evidence and students' self-regulation behaviors that include self-efficacy and goal-setting strategies, challenging prevalent one-time feedback methods \cite{naik2025providing,yu2025using,malik2025scaffolding}. Traditional LLM training methods, such as reinforcement learning from human feedback (RLHF; \citet{ouyang2022training}), can be adapted to consider feedback quality by incorporating a ``zone of proximal development loss.'' However, if LLM training continues to prioritize human satisfaction, it may limit opportunities for critical thinking and deviate from theoretical foundations, highlighting the need for pedagogically grounded systems.

Our study focuses on English-speaking sixth-grade Earth Science learners, and its applicability to other age groups, subjects, and languages needs to be investigated. Additionally, we did not use a randomized controlled trial (RCT) to measure Inquizzitor's impact on learning gains and behaviors. However, we offer a needed, foundational step towards implementing cognitive theoretical constructs for LLM-based adaptive scaffolding in education.

\section*{Ethical Statement}
All research, data collection, and analyses were conducted with approval from Vanderbilt University IRB and the Metro Nashville Public Schools system. All study participants---including students, teachers, and parents---provided informed assent and consent to participate. All student data were anonymized prior to any agent interaction or analysis. Inquizzitor was equipped with explicit prompt instructions to avoid engaging in harmful or toxic discussions with students and was rigorously tested by our research team prior to deployment. Anonymized data are available upon request, in accordance with Vanderbilt University IRB guidelines.

\section*{Acknowledgements}
The research reported here was supported by the Institute of Education Sciences (IES), U.S. Department of Education, through Grant R305C240010; and National Science Foundation (NSF) awards IIS-2327708 and DRL-2112635 as subawards to Vanderbilt University. The opinions expressed are those of the authors and do not represent the views of the Institute of Education Sciences, U.S. Department of Education, or National Science Foundation.

\bibliography{references}

\end{document}